\documentclass{ecai-template/ecai} 

\usepackage{latexsym}
\usepackage{amssymb}
\usepackage{amsmath}
\usepackage{amsthm}
\usepackage{booktabs}
\usepackage{enumitem}
\usepackage{graphicx}
\usepackage{color}
\usepackage[colorlinks=true, linkcolor=blue, citecolor=blue]{hyperref}
\usepackage[table]{xcolor} 

\newcommand{\BibTeX}{B\kern-.05em{\sc i\kern-.025em b}\kern-.08em\TeX}

\begin{document}


\begin{frontmatter}


\paperid{123} 


\title{MapATM: Enhancing HD Map Construction through Actor Trajectory Modeling}


\author[A]{\fnms{Mingyang}~\snm{ Li}\thanks{Corresponding Author. Email: mli170@syr.edu.}\footnote{Equal contribution.}}
\author[B]{\fnms{Brian}~\snm{ Lee}\footnotemark}
\author[A]{\fnms{Rui}~\snm{Zuo}}
\author[B]{\fnms{Brent}~\snm{Bacchus}}
\author[B]{\fnms{Priyantha}~\snm{Mudalige}}
\author[A]{\fnms{Qinru}~\snm{Qiu}} 

\address[A]{Department of Engineering and Computer Science, Syracuse University}
\address[B]{Research and Development, General Motors}


\begin{abstract}
High-definition (HD) mapping tasks, which perform lane detections and predictions, are extremely challenging due to non-ideal conditions such as view occlusions, distant lane visibility, and adverse weather conditions. Those conditions often result in compromised lane detection accuracy and reduced reliability within autonomous driving systems. To address these challenges, we introduce MapATM, a novel deep neural network that effectively leverages historical actor trajectory information to improve lane detection accuracy, where actors refer to moving vehicles. By utilizing actor trajectories as structural priors for road geometry, MapATM achieves substantial performance enhancements, notably increasing AP by 4.6 for lane dividers and mAP by 2.6 on the challenging NuScenes dataset—representing relative improvements of 10.1\% and 6.1\%, respectively, compared to strong baseline methods. Extensive qualitative evaluations further demonstrate MapATM's capability to consistently maintain stable and robust map reconstruction across diverse and complex driving scenarios, underscoring its practical value for autonomous driving applications.
\end{abstract}

\end{frontmatter}


\section{Introduction}
\begin{figure}[!t]
\centering
\includegraphics [width=\columnwidth]{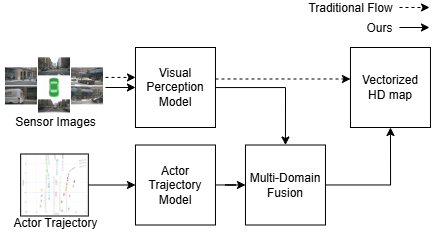}
\caption{\textbf{Architecture of the proposed MapATM framework. }The model combines multi view image features with actor trajectory information through dedicated modules for actor trajectory modeling and multi domain information fusion, resulting in more accurate online HD map construction.}
\label{fig:actor_reasoning}
\vspace{1cm}
\end{figure}

Online High-Definition (HD) mapping is a fundamental task for autonomous driving, providing essential real-time environmental information required for accurate perception, planning, and navigation. Recent advancements in transformer-based vision architectures have significantly advanced online HD mapping capabilities, facilitating critical functionalities in both end-to-end autonomous driving systems\cite{uniad} and geo-fenced HD map-based methods\cite{vectornet}. Approaches utilizing Bird’s-Eye-View (BEV) feature transformations, such as BEVFormer\cite{bevformer} and LSS\cite{lss}, empower query-based detectors like DETR\cite{detr} to effectively detect map elements (e.g., lane dividers), exemplified by frameworks such as MapTR\cite{maptr}, LaneSegNet\cite{lanesegnet}, and StreamMapNet\cite{streammapnet}. State-of-the-art systems such as UniAD\cite{uniad} and VAD\cite{vad} further extend this capability to concurrently detect both actors(e.g., vehicles) and map elements, employing cross-attention mechanisms that enable actor queries to aggregate information from map element queries to enhance actor perception accuracy.\\
Despite these advances, current methods exhibit significant limitations, particularly in dynamic scenarios and under visual uncertainties, such as occlusions, ambiguous lane markings, and adverse weather conditions. Additionally, these methods primarily rely on visual information for detecting map elements, 
often neglecting the rich spatial-temporal context that actors can provide. As a result, detection performance degrades when the line of sight is occluded. Given that vehicles generally adhere to established road rules, their historical trajectories implicitly contain structural information regarding road geometry, including lane centerlines, intersections, and permissible driving paths. For instance, vehicles following consistent lanes or executing predictable maneuvers at intersections provide direct and valuable structural priors that traditional methods do not utilize.\\
Motivated by this observation, we propose Map with Actor Trajectory Modeling (MapATM), a novel architecture designed to explicitly integrate actor trajectory data, which can come from any sensing modality (camera, radar, LiDAR, etc.), into online lane inference tasks. As shown in Fig. \ref{fig:actor_reasoning},  MapATM extends the traditional flow by introducing dedicated modules for actor trajectory modeling and multi-domain information fusion.
During training, MapATM learns the spatial-temporal relationship between actors' trajectories and road geometry, significantly improving map prediction accuracy beyond immediate visual perception.\\

\begin{figure*}[!t]
\centering
\includegraphics [width=0.9\textwidth]{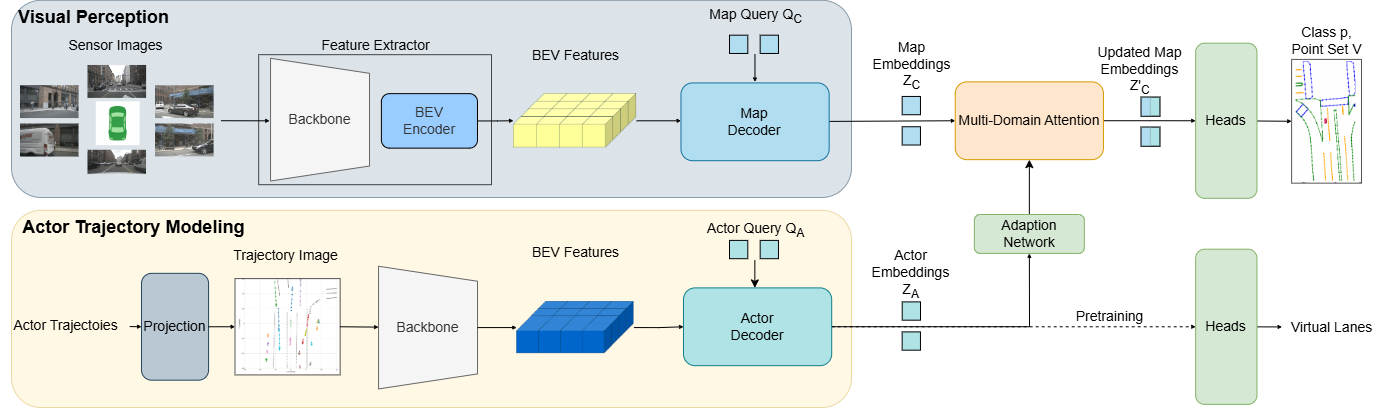}
\caption{\textbf{Overall architecture of MapATM.} Sensor data and transformed actor trajectories are first converted into BEV features. Learnable queries capture map elements (e.g., lanes) and actors (e.g., vehicles). Actor trajectories explicitly encode spatial-temporal map geometry, which is fused with map queries via cross-attention, resulting in enhanced map element inference.}
\label{fig:workflow}
\vspace{0.5cm}
\end{figure*}
Our comprehensive evaluations on the challenging NuScenes dataset demonstrate that MapATM significantly surpasses baseline methods such as VAD, confirming substantial performance improvements and robustness in diverse scenarios. The practical implications of adopting MapATM include enhanced reliability and accuracy in autonomous driving systems, particularly in visually challenging and dynamically complex environments.\\

Our contributions can be summarized as follows:
\begin{itemize}
\item We propose MapATM, an end-to-end approach integrating actor trajectory data into autonomous driving perception to enhance mapping accuracy.
\item MapATM incorporates dedicated modules for information fusion between actor and map elements
, specifically targeting improvements in BEV perception under visual uncertainty scenarios.
\item Comprehensive experimental results confirm that MapATM consistently and outperforms existing perception models by an average of 1.9 mAP in visual occluded scenarios and 2.6 mAP in general scenarios. The results highlight its substantial performance improvements offered by the seamless integrative.
\end{itemize}

\section{Related Work}
\subsection{Semantic Map Construction}
Recent advancements in perception transfer from perspective view images to Bird’s Eye View (BEV) have significantly contributed to this field. LSS \cite{lss} estimates the depth distribution of individual pixels to project perspective view image features into BEV space. On the other hand, BEVFormer \cite{bevformer} projects sampled points from BEV pillars onto camera images, leveraging spatial and temporal attention mechanisms to encode BEV features. In the domain of BEV-based map element prediction, vectorized maps—comprising lane vectors (polygon or polylines)—have gained popularity over rasterized segmentation maps due to their ability to retain structural information. VectorMapNet \cite{vectormapnet} is a pioneering work that introduces an end-to-end vectorized HD map to detect map elements in an autoregressive manner. Meanwhile, MapTR \cite{maptr, maptrv2} models map elements as permutation-equivalent point sets, effectively clarifying their shapes. However, camera-based methods often struggle in scenarios with unclear lane markings and occlusions, particularly in local streets or during peak traffic hours. To address these challenges, MapATM integrates actor information to enhance map representation, reducing uncertainty in road structure estimation.

\subsection{Multi-modal Fusion mapping}
Several works\cite{hdmapnet, vectormapnet, semanticslam} have explored the integration of multiple sensor inputs to enhance mapping performance. MapTRv2 \cite{maptrv2} fuse LiDAR and image information into BEV representations. While this approach improves mapping accuracy, it comes at the cost of significantly reduced inference speed (measured in FPS) due to the computational burden of extracting features from large-scale LiDAR inputs. This limitation restricts their applicability in real-time scenarios. Moreover, cameras and LiDAR operate at different sampling rates, leading to temporal misalignment, which can further degrade real-time perception performance.
Rather than fusing LiDAR and camera streams frame-by-frame, we fuse historical actor trajectories—high-level cues distilled from earlier predictions or simple arithmetic integration—with visual features inside a shared embedding space. Because this merging happens at the information-channel level, it is multi-domain: the same mechanism works whether the inputs are single- or multi-modal. By shifting fusion from raw sensor data to trajectory-guided, channel-level embeddings, we avoid the heavy computation and temporal mis-alignment of conventional multi-modal fusion, achieving higher mapping accuracy with virtually no loss in inference speed.


\section{Method}
\subsection{Problem Formulation}
In this project, there are two inputs: Images from multi view cameras $\mathcal{I_C} = \{I_i\}_{i=1}^{N_I}$, where $N_I$ is the number of cameras, and actor trajectories, $\mathcal{A}=\{A_i\}_{i=1}^{N_A}$, where $N_A$ denotes number of tracking actors. Each actor trajectory $A_i\in\mathbb{R}^{T\times2}$ consist of most recent $T$ 2-dimensional coordinate of the corresponding vehicle under the current ego coordinate system:
\begin{eqnarray}\label{eq:actor}
\mathcal{A} = \left\{ 
A_i = \big[ (x_i^{t - T_i + 1}, y_i^{t - T_i + 1}),\, \ldots,\, (x_i^t, y_i^t) \big] 
\right\}_{i=1}^{N_A}
\end{eqnarray}
We aim to build an online HD map by detecting and classifying map elements, such as lane lines and pedestrian crossings, and representing them using polylines and polygons. Each polyline and polygon is defined as an ordered set of points $\mathcal{V} =\{V_{j}\}_{j=1}^{N_p}$ where $N_p$ is number of points. 

\subsection{Overview}
Fig.\ref{fig:workflow}  illustrates the overall architecture of our proposed method, \textbf{MapATM}. The model takes multi-view images and actor trajectory information as input. The overall processing flow can be divided into two branches, a visual perception branch, which is based on traditional scene representation flow such as VAD\cite{vad}, and an actor trajectory modeling branch, which explicitly models past vehicle motion. In the visual perception branch, a BEV encoder first processes the multi-camera image features, transforming them into a unified egocentric Bird’s-Eye-View (BEV) representation. This BEV feature serves as the input for subsequent perception modules.In the visual perception, a map decoder combines the BEV representation with trained map queries to retrieve embedding vectors for map elements corresponding to potential road elements. 

Meanwhile, the actor trajectory modeling branch processes the trajectory data in parallel. Historical actor trajectories are first projected into the same BEV coordinate, forming a trajectory image in BEV space. This BEV trajectory image is then passed through an actor decoder to extract actor embeddings. To effectively align and integrate information across modalities, an adaptation layer transforms the actor embeddings into a common feature space compatible with the map queries. These actor-informed features are then fused with image-derived BEV features, resulting in an enhanced map representation enriched with actor context. Finally, prediction heads utilize the refined map queries to produce the final lane and topology prediction results, enabling improved online HD map construction informed by both visual and actor-derived information.

\subsection{Details on the Visual Perception}
Similar to prior works, the visual perception in the MapATM performs image-based perception in an end-to-end fashion by learning HD map representations directly from multi-camera inputs. The set of multi-camera images is first passed through a shared backbone to extract image features $\mathcal{F_I} = \{F_{Ii}\}_{i=1}^{N_I}$, where $N_I$ is the number of input camera views.
To fuse both spatial and temporal information across multiple views, we adopt BEVFormer\cite{bevformer}, which transforms the sequential multi-view features into a unified egocentric BEV representation $\mathcal{B}\in\mathbb{R}^{bev_h \times bev_w \times c}$, where $bev_h$ and $bev_w$ are the dimensions of the BEV view and $c$ is the number of channels in the feature vector.

Subsequently, the BEV features are processed by a map decoder with hierarchical map queries $Q_C = Q_{instance}\oplus Q_{point}$ to retrieve point level descriptions for instances of map elements. Here, $Q_{instance}$ and $Q_{point}$ are sets of queries at instance and point levels, and $\oplus$ stands for the broadcast addition. 
The map decoder is implemented using deformable Attention~\cite{deformDETR} model. Its output is a set of retrieved map elements $\mathcal{Z_C} = \{Z_{Ci}\}_{i=1}^{N\times N_p}$ where $N$ is the number of instances of map element and $N_p$ is the number of points associated with each instance.


\subsection{Details of the Actor Trajectory Modeling}
Leveraging a unified model to process each individual actor for feature extraction
presents significant challenges due to variations in their quantities, heterogeneous tracking durations, and diverse spatial distributions. To address this issue, we propose transforming each actor's historical trajectory from raw egocentric 2D position into a Bird's Eye View (BEV) representation. A unified binary  trajectory image $\mathcal{I_A}\in\mathbb{R}^{h \times w}$ is generated that combines the trajectory information all actors, where a "1" indicates that the location is part of the trajectory of an actor in the last $T$ steps. A backbone network is then used to extract trajectory image features $F_A$. This process consolidates diverse actor trajectories into a consistent spatial representation, effectively encoding historical actor movements into a single, structured feature map. \\
Recognizing that the BEV transformation inherently decoupled critical tracking details, particularly trajectory connectivity between positions, which is essential for inferring road geometry, we propose an additional mechanism inspired by autoencoder techniques. Specifically, we leverage deformable attention model and a set of trained actor queries $Q_A$ to encode individual trajectory elements from the trajectory image into actor embeddings $\mathcal{Z}_A=\{Z_{ai}\}_{i=1}^{N'_A}$, where $N'_A$ denotes the number of actor queries:
\begin{eqnarray}\label{eq:ar_autoencoder}
\mathcal{Z}_{A} = DeformAttn(Q_A, F_A)
\end{eqnarray}
These embeddings encapsulate high-level trajectory characteristics, such as direction and velocity, ensuring retention of essential geometry information. 

To train the queries $Q_A$ and the deformable attention, a feed-forward networks (FFNs) subsequently is used to  decode these embeddings to virtual map elements $\hat{Y}_A=\{\hat{y}_{ai}\}_{i=1}^{N'_A}$:
\begin{eqnarray}\label{eq:ar_head}
\hat{Y}_A = FFNs(\mathcal{Z}_{A})
\end{eqnarray}

Rather than using ground-truth lane annotations—which would force a tight one-to-one correspondence between noisy trajectories and precise road geometry—we construct virtual lane edges directly from the trajectories themselves. Specifically, for every tracked actor we shift its historical path perpendicularly to the left and right by half the nominal lane width, generating two offset curves that approximate the lane boundaries surrounding that vehicle.  The resulting target set is $\hat{y}_{ai}=\{\hat{p}_{ai}, \hat{V}_{ai} \}$, where $\hat{p}_{ai}$ gives the classification score and $\hat{V}_{ai}$ is a set of points. The target value of virtual map is $Y_A=\{y_{aj}\}_{j=1}^{2\times N_A}$ where $\{y_{aj}\}=(c_{aj},V_{aj},\Gamma_{aj})$ are target class label, point set and permutation group of point set for virtual lane edge. To be note, the point set $V_{aj}\in\mathbb{R}^{N_p\times2}$ are generated by shifting each actor's trajectory perpendicular to the trajectory direction which is more associated with lane edges. The deformable attention and the FFNs forms an auto-encoder and they are trained together end-to-end.

To keep permutation invariance of point set during the learning process, $\Gamma_{aj}$ is all permutations of the point set $V_{aj}$. Similar to MapTR\cite{maptr}, the loss can be expressed as:
\begin{equation} \label{eq:ar_loss}
\begin{split}
\mathcal{L}_{actor} = \lambda_1\,\mathcal{L}_{focal}(\hat{p}_{\pi({ai})}, c_{ai})
+ \alpha_1\,\mathcal{L}_{distance}(\hat{V}_{\pi({ai})}, V_{ai}) \\
\quad + \beta_1\,\mathcal{L}_{direction}(\Delta\hat{V}_{\pi({ai})}, \Delta V_{ai})
\end{split}
\end{equation}

where $\Delta V$ denotes the vector of consecutive difference of $V$. The loss is a combination of Focal loss\cite{focalloss} $\mathcal{L}_{focal}$, Manhattan Distance loss $\mathcal{L}_{distance}$ and cosine similarity loss $\mathcal{L}_{distance}$ between prediction and groundtruth.
To assign predictions to targets, we adopt the Hungarian algorithm, an efficient method that finds the optimal one-to-one matching by minimizing a predefined cost, here composed of the focal loss and Manhattan distance.
This comprehensive framework accommodates a dynamic number of actors with varying tracking durations, thereby achieving robust and effective extraction of actor information vital for precise lane and topology predictions.

\subsection{Multi-Domain Information Fusion}
Actor trajectories do not necessary align with lane information. There are noises due to natural deviations and irregular driving behaviors, such as drifting, lane weaving, lane merging, or erratic movements like lane straddling and lane splitting. However, despite those noises, actor trajectories still retain an underlying structure reflecting  road topology, providing a useful prior for map element detection. 

To effectively retrieve and utilize the actor-derived prior, we leverage attention mechanism to fuse features from map elements, $\mathcal{Z_C}$, and actor trajectories, $\mathcal{Z_A}$. 
Specifically, we employ map query features $\mathcal{Z}_C$ extracted from camera images, where each query corresponds to potential map elements.
These queries exploit the actor trajectories to resolve the associated uncertainties in road geometry.  This, again, is achieved using deformable attention mechanisms \cite{deformDETR}, which dynamically integrates actor trajectories and visual features to enhance the accuracy and robustness of road structure predictions under diverse scenarios:
\begin{eqnarray}\label{eq:cross_attn}
\mathcal{Z'}= DeformAttn(\mathcal{Z}_C, \mathcal{Z}_A)
\end{eqnarray}
The enhanced feature $\mathcal{Z'}$ then feed to the final FFNs, to decode map elements. Like actor trajectory decoding, each final decoded map element $\hat{y}_{i}=\{\hat{p}_i, \hat{V}_i \}$ includes classification score and points. The groundtruth is $Y=\{y_j\}_{j=1}^{N}$ where $\{y_j\}=(c_j,V_j,\Gamma_j)$ are target class label, point set and permutation group of point set. 
Similarly, the map loss can be formulated as:
\begin{equation} \label{eq:map_loss}
\begin{split}
\mathcal{L}_{map} = \lambda_2\,\mathcal{L}_{focal}(\hat{p}_{\pi(i)}, c_i)
+ \alpha_2\,\mathcal{L}_{distance}(\hat{V}_{\pi(i)}, V_i) \\
\quad + \beta_2\,\mathcal{L}_{direction}(\Delta\hat{V}_{\pi(i)}, \Delta V_i)
\end{split}
\end{equation}

\begin{table*}[t]
\caption{Comparison on nuScenes val set. The APs of VAD are taken from the official checkpoints for all dataset, and fine tuned for complex dataset.}
\vspace{10pt}
\label{tb:res}
\centering
\begin{tabular}{l c c c c c c c}
\toprule
Method & Dataset  & $AP_{ped}$ & $AP_{divider}$ & $AP_{boundary}$ & mAP\\
\midrule
VAD & general    & 37.7 & 45.2 & 44.3 & 42.4\\
\rowcolor{gray!20}
MapATM & general  & \textbf{41.1} & \textbf{49.8} & 44.1 & \textbf{45.0}\\
\midrule
VAD & complex    & 34.1 & 45.1 & 47.1 & 42.1\\
\rowcolor{gray!20}
MapATM & complex  & \textbf{35.1} & \textbf{49.7} & 47.0 & \textbf{44.0}\\

\bottomrule
\end{tabular}
\end{table*}

\begin{figure*}[h]
\centering
\includegraphics[scale=0.4]{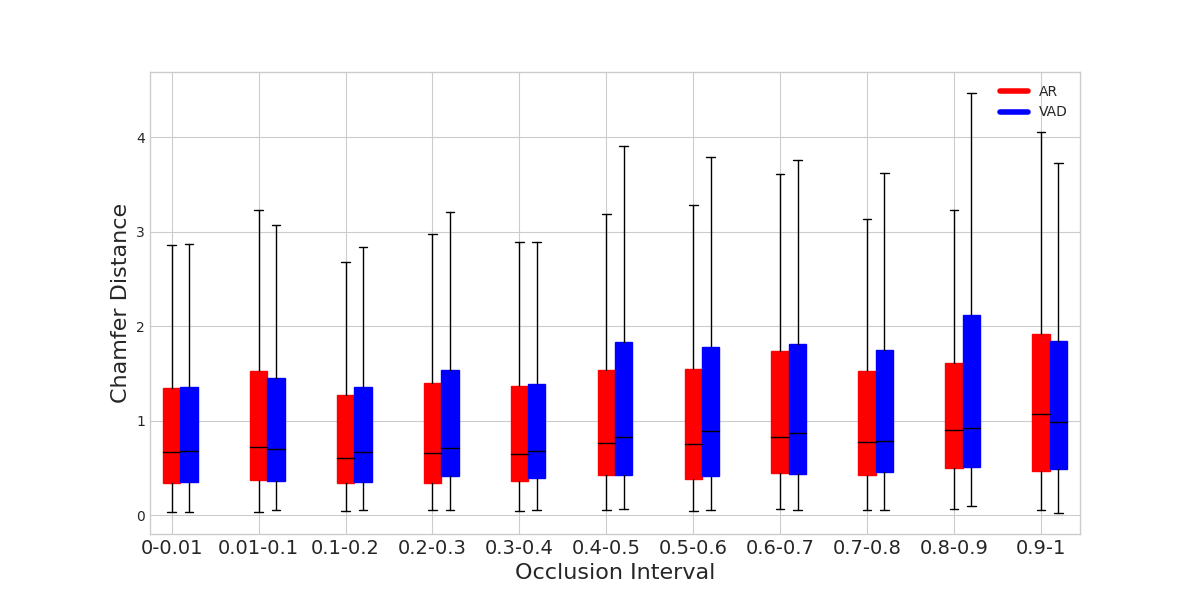}
\caption{Comparison of Distance under different occlusion interval}
\label{fig:distance}
\end{figure*}

\section{Experiments}
\subsection{Implementation Details}
\textbf{Dataset and Metric.} We conduct our experiments on the nuScenes dataset~\cite{nuscenes}, a large-scale public benchmark for autonomous driving. It consists of 1000 scenes collected from Boston (left-hand traffic) and Singapore (right-hand traffic), with each scene spanning 20 seconds and sampled at 2 Hz. In our setup, we use six camera images forming a 360\textdegree{} horizontal view as input, along with actor trajectory information obtained from LiDAR sensors. Given the sparsity of actor annotations in nuScenes, we curate a more complex subset of the dataset. This is selected based on lane occlusion rates and the number of available actor trajectories, retaining 48.26\% and 45.89\% of the original training and evaluation sets, respectively. We refer to the selected dataset as "complex" dataset as opposed to the original "general" dataset.\\
The map prediction task focuses on three types of map elements: lane dividers, road boundaries, and pedestrian crossings. The perception range in BEV space is defined as [-15.0m, 15.0m] along the x-axis and [-30.0m, 30.0m] along the y-axis, centered at the ego vehicle. For evaluation, we employ the Chamfer Distance to measure the geometric discrepancy between predicted and ground-truth polylines. Prediction quality is assessed using Average Precision (AP) under multiple Chamfer Distance thresholds {0.5m, 1.0m, 1.5m} when prediction score is greater than predefined threshold(e.g. 0.7), along with their average mAP, offering a robust measure of both coarse and fine-grained map accuracy.\\
\textbf{Baseline} We use VAD~\cite{vad} as our baseline, which is a vectorized scene-representation network that converts multi-view camera images into BEV features and jointly detects actors and HD-map elements.\\
\textbf{Training details} We build our method on top of VAD-tiny~\cite{vad}, an autonomous driving framework that provides a perception module serving as the foundation for our image-based perception. The Actor Trajectory Modeling using ResNet~\cite{resnet} and FPN~\cite{fpn} as the backbone for trajectory image processing. 
We initialize our model and baseline with the same official VAD checkpoints trained for 60 epochs and then fine-tune them on the constructed actor-focused complex subset. MapATM is trained with the same hyperparameter as the VAD to ensure a fair comparison. All experiments are conducted using a single NVIDIA Tesla A100 GPU.

\subsection{Result}
In Table~\ref{tb:res}, we compare the performance of our proposed model, MapATM, with the VAD baseline on both the "general" and "complex" datasets. The integration of actor trajectories leads to clear improvements in map element detection. On the general dataset, MapATM achieves 4.6 higher AP for lane dividers and 2.6 higher mAP. On the complex dataset, it shows a 3.4 AP gain for lane dividers and a 1.9 increase in mAP.
These results highlight the effectiveness of incorporating actor trajectory information, particularly for elements like lane dividers that closely correlate with vehicle movement.

The precision improvement that the MapATM gains in lane divider predictions validate our hypothesis that actor trajectory serve as reliable priors for inferring road structure. 
In contrast, no gain is observed for road boundaries, and improvements for pedestrian crossings are relatively minor, as these elements exhibit weaker correlations with vehicle trajectories.
Consequently, our model implicitly deprioritizes such elements, focusing on those where actor information provides more significant guidance, without adversely affecting the overall performance.

For a more detailed analysis, we examine MapATM's performance under varying occlusion levels. Occlusion rates are calculated as the ratio of the occluded length of a map element to its total length within a scene. As shown in Figure~\ref{fig:distance}, MapATM consistently identifies map elements that has lower Chamfer Distance to the ground truth map elements compared to the baseline. The observation confirms the enhanced detection accuracy through actor-informed reasoning.
We also observed a slight performance drop in scenarios with complete visual occlusion. This is attributed to the dominant role of trajectory information in the absence of visual cues, which can lead to increased false positives due to noise in the trajectory data. Despite this, MapATM maintains robust performance in most challenging scenarios, demonstrating its practical utility in real-world autonomous driving environments.



\begin{table}[t]
\caption{Ablations about multi-modality Fusion}
\vspace{10pt}
\label{tb:prefusion}
\centering
\resizebox{\columnwidth}{!}{%
\begin{tabular}{l c c c c}
\toprule
Modeling method &  $AP_{ped}$ & $AP_{divider}$ & $AP_{boundary}$ & mAP\\
\midrule
VAD & 34.1 & 45.1 & 47.1 & 42.1\\
\midrule
Pre-Fusion   & 33.1 & 45.7& 43.8 & 40.9\\
Pre-Fusion + Projection  & 33.3 & 44.8 & 45.0 & 41.0 \\
\rowcolor{gray!20}
In-Fusion(Our) & \textbf{35.1} & \textbf{49.7} & \textbf{47.0} & \textbf{44.0}\\
\bottomrule
\end{tabular}
}
\end{table}

\begin{table}[t]
\caption{Ablations about Fusion Methods}
\vspace{10pt}
\label{tb:fusion_method}
\centering
\resizebox{\columnwidth}{!}{%
\begin{tabular}{l c c c c}
\toprule
Fusion method &  $AP_{ped}$ & $AP_{divider}$ & $AP_{boundary}$ & mAP\\
\midrule
Layer-wise attention    & 30.5 & 48.4& 46.8 & 41.9\\
\rowcolor{gray!20}
DeformDETR\cite{deformDETR} & \textbf{35.1} & \textbf{49.7} & \textbf{47.0} & \textbf{44.0}\\
\bottomrule
\end{tabular}
}
\end{table}

\begin{table}[t]
\caption{Ablations about modeling method of Actor Trajectory Modeling}
\vspace{10pt}
\label{tb:modeling}
\centering
\resizebox{\columnwidth}{!}{%
\begin{tabular}{l c c c c}
\toprule
Modeling method &  $AP_{ped}$ & $AP_{divider}$ & $AP_{boundary}$ & mAP\\
\midrule
Backbone    & 33.5 & 47.8& 47.0 & 42.7\\
Backbone + Encoder & 32.4 & 47.6 & 45.9 & 41.9 \\
\rowcolor{gray!20}
Backbone + Actor query & \textbf{35.1} & \textbf{49.7} & \textbf{47.0} & \textbf{44.0}\\
\bottomrule
\end{tabular}
}
\end{table}

\begin{table}[t]
\caption{Ablations about actor quries of Actor Trajectory Modeling}
\vspace{10pt}
\label{tb:autoencoder}
\centering
\resizebox{\columnwidth}{!}{%
\begin{tabular}{l c c c c}
\toprule
Pretraining Output &  $AP_{ped}$ & $AP_{divider}$ & $AP_{boundary}$ & mAP\\
\midrule
Virtual centerline & 31.4 & 44.5& 46.3 & 40.7\\
\rowcolor{gray!20}
Virtual lane edge & \textbf{35.1} & \textbf{49.7} & \textbf{47.0} & \textbf{44.0}\\
\bottomrule
\end{tabular}
}
\end{table}

\begin{figure*}[!t]
\centering
\includegraphics [width=0.95\textwidth]{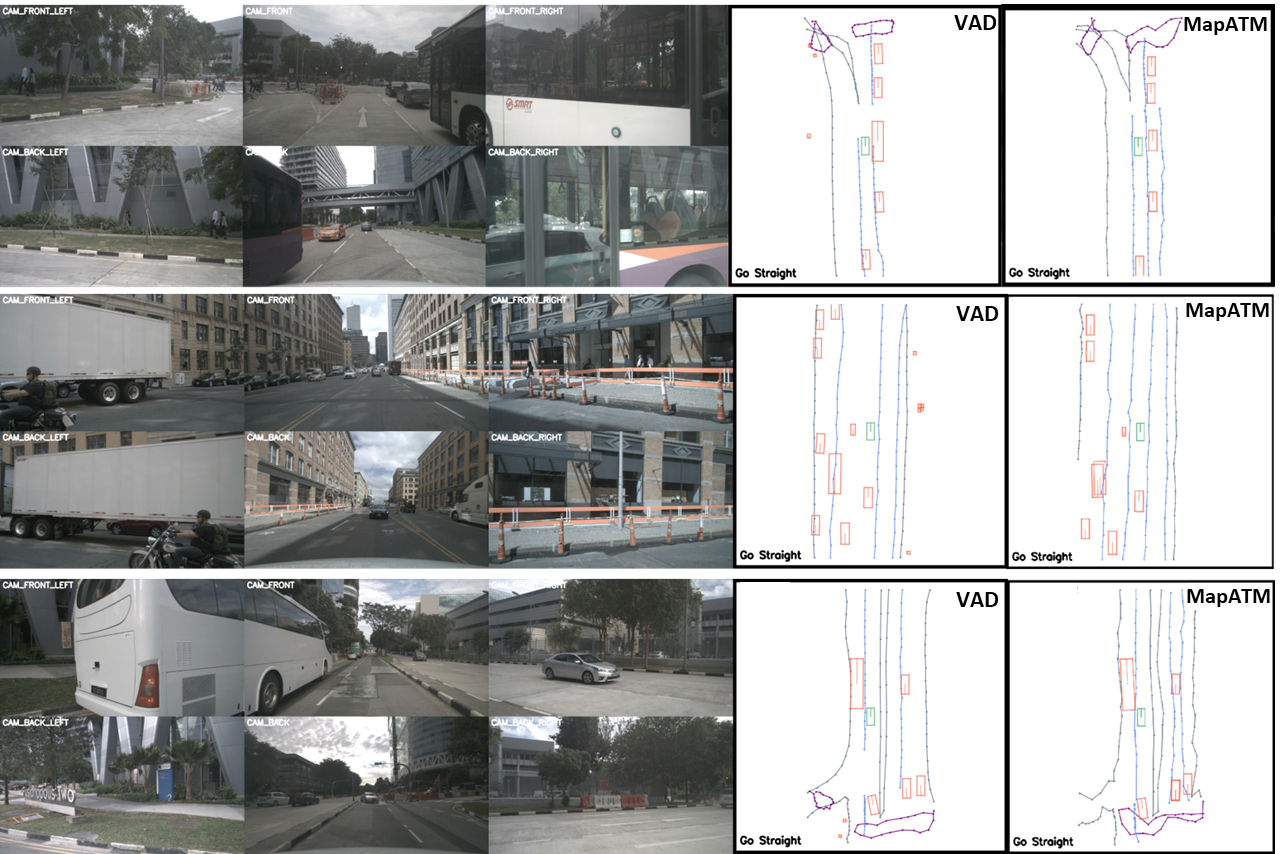}
\caption{Qualitative results under occlusion. The left vectorized HD map is from VAD, the right is MapATM, and the green and red boxes represent the ego car and surrounding cars respectively.}
\label{fig:qualitative}
\end{figure*}

\subsection{Ablation Studies}
\textbf{Multi-domain Information Fusion} We explore different strategies for fusing visual map features and actor trajectory features. Fusion is tested at different stages of the architecture. In the pre-fusion setting, actor features are combined with BEV features directly after the BEV encoder, and the fused representation is then passed to the map decoder. Two variants are considered: the first one alters the BEV encoder output dimension to accommodate concatenation, and the second one (pre-fusion+projection) projects the concatenated output back to the original BEV dimension. The comparison results are shown in Tab.\ref{tb:prefusion}, 
We can see that the pre-fusion not only fails to improve but actually degrades performance. This degradation is likely due to direct, position-wise addition of noisy actor features into BEV space, which disrupts visual features that are unrelated to actor trajectories. 

In contrast, our proposed in-fusion technique first extracts map features from BEV and then selectively incorporates actor features using attention mechanisms. This targeted fusion allows the model to build meaningful relationships between actors and map elements, leading to a performance boost.\\
\textbf{Fusion Method}. We ablate different multi-domain fusion methods, comparing standard layer-wise cross-attention with Deformable DETR~\cite{deformDETR}-based attention. As shown in Tab.~\ref{tb:fusion_method}, Deformable Attention achieves superior results. This indicates that letting candidate map features query relevant actor context through learned attention provides more effective feature enhancement and uncertainty reduction.\\
\textbf{Effectiveness of Actor Trajectory Modeling Modeling} We further validate the effectiveness of our actor trajectory modeling module. The results are shown in Tab.\ref{tb:modeling}. Compared with simpler alternatives—such as directly using trajectory image features or feeding them into a Transformer encoder—our actor query design better captures high-level trajectory semantics like direction and velocity. This helps infer and extend map elements beyond the visible field, yielding a 1.3 mAP improvement and confirming the value of structured actor trajectory modeling.\\
\textbf{Effectiveness of Virtual Lane Pretraining} Our actor trajectory modeling module employs learnable actor queries that encode trajectory image features via an autoencoder-style setup. We evaluate two pretraining strategies in Table~\ref{tb:autoencoder}: predicting virtual centerlines and predicting virtual lane edges. Results show that pretraining with lane edge prediction, which aligns closely with our final task, leads to superior performance. This suggests the model more effectively learns geometric patterns relevant for lane detection.
\subsection{Qualitative Visualization} We present qualitative comparisons of vectorized HD map predictions between our method and the VAD baseline. As shown in Figure~\ref{fig:qualitative}, MapATM successfully detects lane dividers even in visually occluded regions. For example, it accurately predicts the lane divider despite the presence of obstructing objects, demonstrating its robustness and enhanced reasoning capabilities.

\section{Conclusions}
In conclusion, we have proposed a model with explicit Actor Trajectory Modeling. This model utilize actor trajectories and projects them into the BEV image space, integrating them into map embeddings. This approach significantly improves mapping performance, particularly in enhancing lane divider detection, with minimal impact on road boundary and pedestrian crossing detection. Looking ahead to future work, we aim to advance the Actor Trajectory Modeling model to predict centerlines. This will contribute to topology graph inference and enhance the overall capabilities of our automated system.

\bibliography{ecai-template/mybibfile}

\end{document}